\documentclass[letterpaper, 10 pt, conference]{ieeeconf}  % Comment this line out if you need a4paper

\IEEEoverridecommandlockouts                              % This command is only needed if 
\usepackage{graphics} % for pdf, bitmapped graphics files
\usepackage{epsfig} % for postscript graphics files
\usepackage{cite}
\usepackage{times} % assumes new font selection scheme installed
\usepackage{amsmath} % assumes amsmath package installed
\usepackage{amssymb}  % assumes amsmath package installed
\usepackage{graphicx}
\usepackage{mathrsfs}   
\usepackage{caption}
\usepackage{subcaption}% Needed to meet printer requirements.
\usepackage{url}
\usepackage{color}
\usepackage{algorithm}
\usepackage{algorithmic}

\newtheorem{Definition}{Definition}

\newtheorem{Assumption}{Assumption}

\newtheorem{Problem}{Problem}
\newtheorem{theorem}{Theorem}
\newtheorem{remark}[theorem]{Remark}

\newtheorem{proposition}{Proposition}

\title{\LARGE \bf
Safe Navigation in Human Occupied Environments Using Sampling and Control Barrier Functions
}

\author{Keyvan Majd$^{1}$, Shakiba Yaghoubi$^{2}$, Tomoya Yamaguchi$^{2}$, Bardh Hoxha$^{2}$, Danil Prokhorov$^{2}$, Georgios Fainekos$^{1}$
\thanks{$^{1}$K. Majd and G. Fainekos ({\tt\footnotesize
 \{majd,fainekos\}@asu.edu}) are with CIDSE, Arizona State University, Tempe, AZ, USA.}%
\thanks{This research was partially funded by NSF OIA 1936997}
\thanks{$^{2}$S. Yaghoubi, T. Yamaguchi, D. Prokhorov, and B. Hoxha ({\tt \footnotesize
<first\_name.last\_name>@toyota.com}) are with the Toyota Research Institute of North America, Ann Arbor, MI, USA.}}

\begin{document}

\maketitle
\thispagestyle{empty}
\pagestyle{empty}

%%%%%%%%%%%%%%%%%%%%%%%%%%%%%%%%%%%%%%%%%%%%%%%%%%%%%%%%%%%%%%%%%%%%%%%%%%%%%%%%
\begin{abstract}
Sampling-based methods such as Rapidly-exploring Random Trees (RRTs) have been widely used for generating motion paths for autonomous mobile systems. In this work, we extend time-based RRTs with Control Barrier Functions (CBFs) to generate, safe motion plans in dynamic environments with many pedestrians. Our framework is based upon a human motion prediction model which is well suited for indoor narrow environments. We demonstrate our approach on a high-fidelity model of the Toyota Human Support Robot navigating in narrow corridors. We show in three scenarios that our proposed online method can navigate safely in the presence of moving agents with unknown dynamics.
\end{abstract}
%%%%%%%%%%%%%%%%%%%%%%%%%%%%%%%%%%%%%%%%%%%%%%%%%%%%%%%%%%%%%%%%%%%%%%%%%%%%%%%%
\section{INTRODUCTION}

\begin{figure}[t]
	\centering
	\includegraphics[scale=0.27]{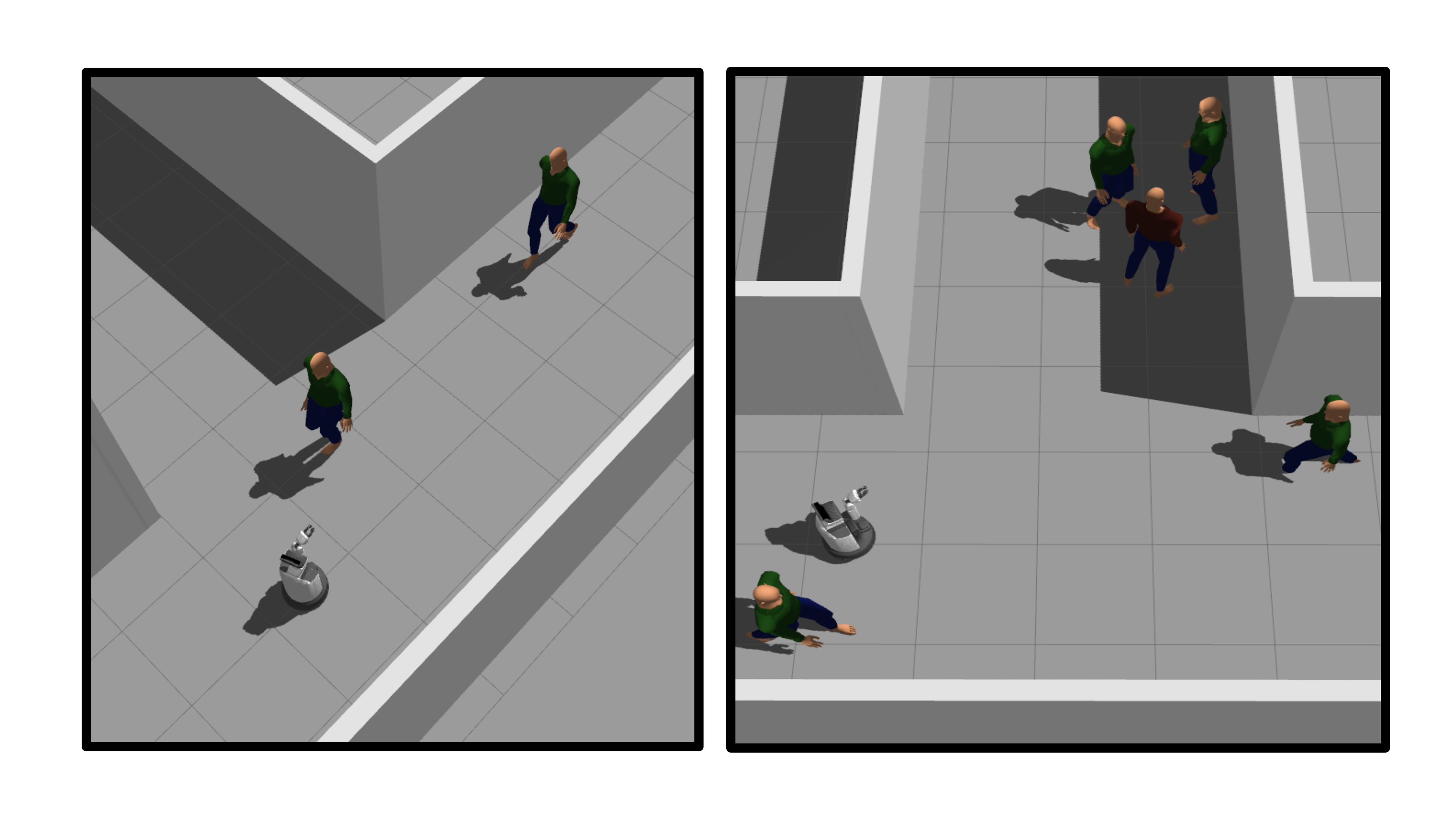}\vspace{-0pt}
	\caption{Simulation scenarios for the robot navigation in narrow corridors using the proposed CBF-TB-RRT framework.}
	\label{fig:gazebo}
\end{figure}

The integration of autonomous mobile systems (AMS) in human environments requires rigorous techniques to ensure safety.
A wide variety of AMS have demonstrated utility in environments such as airports, hospitals etc \cite{deemyad2020chassis, khalilisenobari2021impact}.  As part of this process, an AMS needs to generate a motion plan that makes progress towards the goal while maintains a safe distance to humans in the environment.  Due to the unpredictability and complexity of human motion, safe navigation in such environments still remains a challenge\cite{dadvar2021contemporary}. In this paper, we present a method for start-to-goal motion planning \cite{petti2005safe, habibian2021design}. In particular,  we focus on navigation through narrow corridors, while satisfying safety requirements in the presence of unknown dynamic obstacles (see Fig. \ref{fig:gazebo}). \

The motion planning and control synthesis problem has been studied extensively in the past \cite{sucan2012open,kleinbort2018probabilistic}. 
Sampling-based algorithms such as RRT and PRM have been widely used for path exploration\cite{hsu2002randomized,tuncali2019rapidly}. The simplicity and practical efficiency of these methods have been proved for motion planning in complex environments under kinodynamic constraints \cite{palmieri2016rrt,hernandez2019lazy}. Another set of works encode the planning problem into mixed-integer quadratic programs \cite{mellinger2012mixed} or nonlinear model predictive control frameworks \cite{nishimura2020risk}. These methods solve a low-level planning problem and they could be considered as local repair solutions to the general motion planning problem. In \cite{shoukry2016scalable}, the planning problem for reach-avoid requirements is posed as a satisfiability problem over linear constraints. The approach scales linearly with the number of agents; however, it is limited to static environments and linear system dynamics. \

Control Barrier Function (CBF) methods, presented in \cite{yaghoubi2020training,ames2019control}, have been used for control synthesis for safety-critical systems. They may be combined with Lyapunov Functions to form a Quadratic Program (QP) that produces safe and stable control in a feedback control setting \cite{yang2019self,cheng2019end, lindemann2019decentralized, lindemann2018control}. While CBF-based feedback controllers guarantee the generated control to be safe, they may become infeasible or get trapped in local minima. To address this issue, in \cite{yang2019sampling}, the CBF-based navigation synthesis is combined with a sampling-based planner to ensure collision-free trajectories. The main drawback of this method is that it assumes the dynamics of moving agents to be known. Therefore, in the presence of obstacles with unknown dynamics, the implementation of this method is not applicable in real-time. TB-RRTs \cite{wang2020eb, fulgenzi2010risk}  have been developed for dynamic environments, where planning occurs in real-time. There is a necessary compromise between the expansion of the tree (exploration) and the timing requirements in order to return a feasible path for the next cycle. In \cite{fulgenzi2010risk}, the authors present time-based Risk-RRTs which utilize an algorithm based on Gaussian processes to predict the motion of other agents in the environment. Different from \cite{fulgenzi2010risk}, our method integrates the dynamics of the vehicle with CBF to extend the TB-RRT method resulting in a low-level controller that satisfies safety requirements. \

This paper proposes a CBF-based TB-RRT (CBF-TB-RRT) to address the problem of safety guaranteed motion planning in environments with unknown but predictable dynamic obstacles. Unlike \cite{yang2019sampling}, TB-RRT allows us to update our knowledge from the environment in real time while exploring the environment for a safe path that takes us to the defined goal through CBF-based navigation. We use the planning-based prediction model proposed in \cite{rudenko2018joint} that accounts for local interactions based on Joint Sampling Markov Decision Process (JS-MDP). JS-MDP generates a map that indicates likelihood of human presence. We employ this method to predict the regions occupied by dynamic obstacles in the future and avoid the regions with high probability of occupancy using CBF-based constraints.\

The key contributions of this paper are as follows. First, we utilize TB-RRT in conjunction with CBF to produce motion plans that robustly satisfy collision requirements in real time (Sec. \ref{SEC:CBF-RRT}).
Second, we derive CBF safety constraints assuming the dynamics of moving objects are not known. To this end, we formulate the CBF constraints to avoid the predicted obstacle regions and generate controls to minimize the risk of collision through staying in the predicted safe region (Sec. \ref{SEC:safe-steering} and \ref{SEC:implement-details}). Finally, in three scenarios (see Fig. \ref{fig:gazebo}), we demonstrate the efficiency of our algorithm in generating safe trajectories through narrow corridors tested on a high-fidelity Toyota Human Support Robot (HSR) \cite{yamamoto2019development} model in the Gazebo simulator (Sec. \ref{sim-sec}).\

\section{PRELIMINARIES}

\subsection{Control Barrier Function (CBF)}
We consider a robot motion model with the following nonlinear affine control dynamics
\begin{align}
\label{sys_dyn1}
 \dot{\mathbf{x}} (t)=\mathbf{f}(\mathbf{x}(t))+\mathbf{g}(\mathbf{x}(t))\mathbf{u}(t).
\end{align}
Here, $\mathbf{x}(t)\in \mathcal{X} \subseteq \mathbb{R}^{n }$ denotes the state of robot at time $t$,  $\mathbf{u}(t)\in \mathcal{U} \subseteq \mathbb{R}^{m}$ is the control input, and $\mathbf{f}:\mathbb{R}^{n } \rightarrow \mathbb{R}^{n }$ and $\mathbf{g}:\mathbb{R}^{n } \rightarrow \mathbb{R}^{n \times m}$ are locally Lipschitz functions. \

A function $\alpha: \mathbb{R} \rightarrow \mathbb{R}$ is an extended class $\mathcal{K}$ function iff it is strictly increasing and $\alpha(0) = 0$ \cite{ames2019control}.
A set $\mathcal{C} \subseteq \mathbb{R}^{n }$ is forward invariant w.r.t the system (\ref{sys_dyn1}) iff  for every $\mathbf{x}(0) \in \mathcal{C}$, its solution satisfies $\mathbf{x}(t)\in \mathcal{C}$ for all $t \geq 0$ \cite{blanchini1999set}.

% \begin{Definition}[Barrier Function]\label{defBF}
% Let $h:X \rightarrow \mathbb{R}$ be a continuously differentiable function, $C= \ \{ x \in X | h(x) \geq 0 \}$, and $\alpha$ be a locally Lipschitz extended class $\mathcal{K}$ function.
%  $h$ is a \emph{barrier function} iff for all $x\in C$
% \begin{equation}\label{BF}
% \dot{h}(x)\geq -\alpha(h(x))
% \end{equation}
% \end{Definition}

% \begin{Lemma}[\cite{glotfelter2017nonsmooth}]\label{lemma1}
%  If $h$ is a barrier function for $C$, and $\alpha$ is as defined in Def.~\ref{defBF} then $C$ is a forward invariant set.
% \end{Lemma}

\begin{Definition}[Control Barrier Function \cite{ames2019control}]
A continuously differentiable function $h(\mathbf{x})$ is a Control Barrier Function (CBF) for the system (\ref{sys_dyn1}), if there exists a class $\mathcal{K}$ function $\alpha$ s.t. $\forall \mathbf{x}\in \mathcal{C}$ :
\begin{equation}\label{CBF}
\sup_{u\in U}\big(L_f h(\mathbf{x}) +L_g h(\mathbf{x}) u +\alpha(h(\mathbf{x}))\big)\geq 0 
\end{equation}
where $L_f h(\mathbf{x}) = \frac{\partial h}{\partial \mathbf{x}}^\top f(\mathbf{x}), L_g h(\mathbf{x})= \frac{\partial h}{\partial \mathbf{x}}^\top g(\mathbf{x})$ are the first order Lie derivatives of the system.
\end{Definition}

Any Lipschitz continuous controller $\mathbf{u} \in K_{cbf}(\mathbf{x}) = \{\mathbf{u}\in \mathcal{U}\;|\;L_f h(\mathbf{x}) +L_g h(\mathbf{x}) \mathbf{u} +\alpha(h(\mathbf{x}))\geq 0\}$ results in a forward invariant set $\mathcal{C}$ for the system of Eq. (\ref{sys_dyn1}).
\subsection{Time-Based Rapidly-exploring Random Tree (TB-RRT)}
\begin{figure}[t]
	\centering
	\includegraphics[scale=0.4]{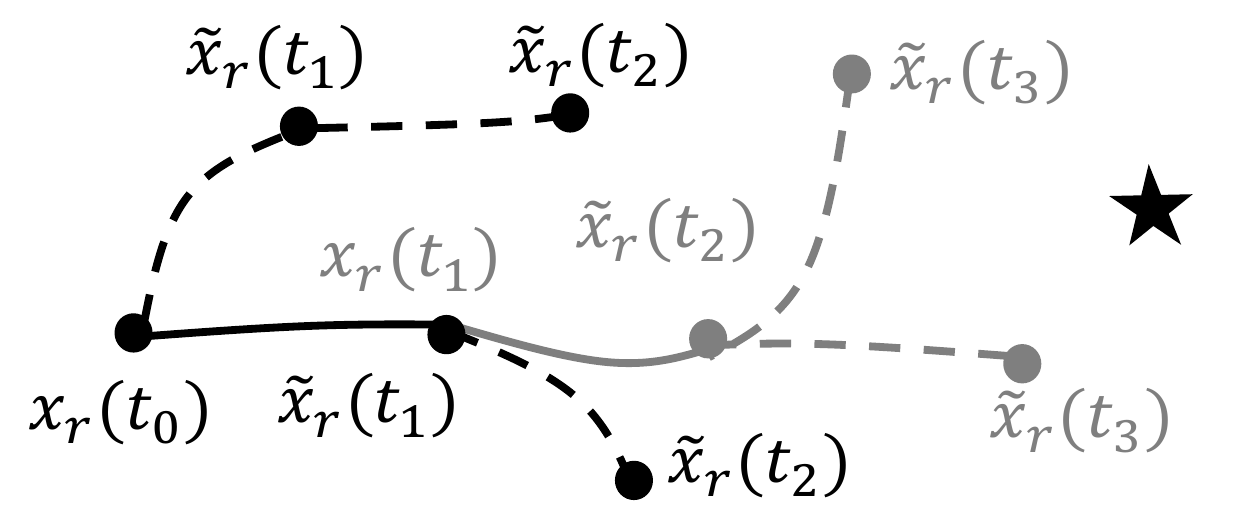}\vspace{-3pt}
	\caption{The black and light grey solid lines connecting the dots are the executed trajectories by the robot in the first ($t_1$) and the second ($t_2$) cycles, respectively. The dashed lines are the generated edges by TB-RRT. The black and grey trees are expanded at the first and second iterations toward goal (shown with star), respectively.}
	\label{fig:time-based-rrt}\vspace{-5pt}
\end{figure}

TB-RRT algorithms, such as those in \cite{wang2020eb} and \cite{fulgenzi2010risk}, provide a real-time solution to the motion planning problem in a dynamic environment. The basic structure of TB-RRT is shown in Fig. \ref{fig:time-based-rrt}. These algorithms generate partial paths iteratively and expand a tree in real-time by limiting the planning time or restricting number of nodes in each cycle until reaching the goal set. To this end, the robot anticipates the behavior of obstacles in real-time and expands a time-stamped tree in a bounded horizon manner. After each planning cycle, the algorithm selects a vertex based on some cost and extracts a partial path from the current position to the selected vertex. While the robot is moving along the selected partial path, the algorithm continues expanding the tree in the next planning cycle. The tree is expanded from the end of previous generated path given the updated information received from the environment. Then, a new partial path is passed to the robot for execution. This cycle continues until reaching the goal. 

%%%%%%%%%%%%%%%%%%%%%%%%%%%%%%%%%%%%%%%%
%%%%%%%%%%%%%%%%%%%%%%%%%%%%%%%%%%%%%%%%
\section{PROBLEM FORMULATION}

Assume that $\zeta: \mathcal{X}\rightarrow \mathbb{R}_-$ is a function that defines the goal set of system (\ref{sys_dyn1}) as follows
\begin{equation}\label{G_set1}
    \mathcal{X}_{g} = \{\mathbf{x}\in \mathcal{X}\;|\;\zeta(\mathbf{x})\leq 0 \}.
\end{equation}
We will also denote the obstacle space as $\mathcal{X}_{o}\subset \mathcal{X}$. In a dynamic environment, the obstacle space is time-varying, so it is denoted with $\mathcal{X}_{o}(t)$. At each time $t$, we denote the safe region with $\mathcal{X}_{s}(t)\subseteq\mathcal{X}\backslash\mathcal{X}_{o}(t)$. We define the safe region using a function $h:\mathcal{X}\rightarrow\mathbb{R}_+$\ as follows
\begin{align}\label{safe-set}
    \mathcal{X}_{s}(t) = \{\mathbf{x}\in \mathcal{X}\;|\;h(\mathbf{x}(t))\geq 0 \}.
\end{align}
%At each time $t$, we denote the safe region with $\mathcal{X}_{s}(t)\subseteq\mathcal{X}\times\mathcal{X}_{o}(t)$. We define the safe region for an augmented state $\mathbf{x}_e(t)=[\mathbf{x}(t),\mathbf{x}_o(t)]^T$ using a function $h:\mathcal{X}\times\mathcal{X}_{o}(t)\rightarrow\mathbb{R}_+$\ as follows
%\begin{align}\label{safe-set}
%    \mathcal{X}_{s}(t) = \{\mathbf{x}_e\in \mathcal{X}\times\mathcal{X}_{o}(t)\;| h(\mathbf{x}_e(t))\geq 0 \}.
%\end{align}
We assume that the dynamics of moving obstacles are not known in advance and that an accurate prediction model exists.
\begin{Assumption} \label{assum1}
Future human motion is accurately represented by a predictor that provides a probability distribution over the human's future positions for a given finite horizon.
\end{Assumption}

It is assumed that the state of static obstacles are given. As the robot navigates in the environment, it senses the trajectory of moving objects $d=1,\cdots,D,$ with states $\mathbf{x}^d_o(t)\in\mathcal{X}^d_o(t)\subset\mathcal{X}_o(t)$. The robot records the set of all observed trajectories in $\mathcal{T}=\{\mathcal{T}_{d=1,\cdots,D}\}$, where $\mathcal{T}_{d}=\langle \mathbf{x}^d_o(t)\rangle$ stores the observed trajectory sequence of dynamic agent $d$ for time $t\in \{i T_s\;|\;i=0,1,\cdots,l_d\}$, where $T_s$ is the sampling time and $l_d$ is the last observed sample of obstacle $d$. Given the observed tracklet $\mathcal{T}_d$ of a moving agent $d$, we assume there exists a prediction module which provides level-sets $\hat{\mathcal{X}}^{d}_{o}(t)$ that predict the future occupied region by the moving obstacles.\

%To predict the future trajectory of the obstacles, a prediction model can be used to predict the motion of multiple agents jointly for a long-term horizon \cite{rudenko2020human}. We extract {\color{red} level sets} from the aforementioned prediction model and predict safe sets for a prediction horizon $t_h$ for dynamic obstacles as follows\
%\begin{align}\label{safe-set2}
%    \hat{\mathcal{X}}_{s}(\tau)\!=\! \{\mathbf{x}\!\in\! \mathcal{X}\!\times\! \hat{\mathcal{X}}_{o}(\tau)\!\;| h(\hat{\mathbf{x}}_e(\tau))\geq 0 \},\tau\!\in\![t~t+t_h].
%\end{align}

In this paper, we aim to compute a trajectory $\sigma: [0,T]\rightarrow\mathcal{X}$, where $T$ is a finite time, s.t. $\sigma(0)=\mathbf{x}(0)$, and $\sigma(T)\in \mathcal{X}_{g}$ while staying in the safe region for all $t\in[0,T]$, i.e., $\sigma(t)\in \mathcal{X}_s(t)$. Hence, our problem is formalized as follows,\
\begin{Problem}\label{problem1}
Find an admissible and bounded control input $\mathbf{u}:[0~T]\rightarrow\mathcal{U}$ for the system (\ref{sys_dyn1}) to obtain an admissible trajectory $\sigma(t)\in\mathcal{X}$, s.t.
 $\sigma(0)=\mathbf{x}(0)$, $\sigma(T)\in \mathcal{X}_{g},$ and $\sigma(t)=\mathbf{x}(t)\in \mathcal{X}_{s}(t)$,  $\forall t\in [0~T]$. 
\end{Problem}
%%%%%%%%%%%%%%%%%%%%%%%%%%%%%%%%%%%%%%%%
%\subsection{Obstacle Space}
%This work assumes while the robot navigates in the environment, it receives position information on humans in the scene and updates it at discrete time, with time intervals $\Delta t_{o}$. While the robot is recording the real-time trajectory of humans, it generates {\color{red}probabilistic} predictions from the human motion for a time horizon $T_h$ in the future. The predictions are then extracted in the from of level sets \ 
%\begin{align}
%    \mathcal{X}&_{o}(t+k\Delta t_{o},r^k_{o})=:\\ \nonumber
%    &\bigg\{ \mathbf{x}_o \in \mathcal{X}_{o}\;|\; ([x ,y ]^T- \mathbf{x}_g)^2 - r_g^2 \leq 0 \bigg\}
%\end{align}
%%%%%%%%%%%%%%%%%%%%%%%%%%%%%%%%%%%%%%%%
%%%%%%%%%%%%%%%%%%%%%%%%%%%%%%%%%%%%%%%%
\section{CBF-BASED TB-RRT (CBF-TB-RRT) ALGORITHM} 
In this section, we detail our CBF-based TB-RRT (CBF-TB-RRT) algorithm for motion planning. Algorithm \ref{CBF-alg} presents the motion planner and Alg. \ref{grow-alg} implements CBF-based tree expansion.\

\subsection{TB-RRT Review}\label{SEC:CBF-RRT}
In a time-based RRT, the information stored in each vertex can be defined with a $3$-tuple $(\mathbf{x},z,t)$, where $\mathbf{x}\in\mathcal{X}$ is the state of robot, $z$ is the cost of vertex, and $t$ is the vertex timestamp.\
At a given time $t_0$, $\textsc{Robot-Pose}(\cdot)$ returns the state $\mathbf{x}$ of the robot at real physical time $t_0$, $\textsc{Obs-Pose}(\cdot)$ returns the positions of the obstacles that are later stored in the set of all observed tracks $\mathcal{T}$, and $\textsc{Execute-Control}(\cdot)$ applies the obtained control to the robot.

In this algorithm, robot performs the planning task by iteratively expanding a tree at each time sample $T_s\in\mathbb{R}$. The choice of $T_s$ depends on the nature of the environment and the robot's control execution time. At each time sample, the obstacle predictor query $\textsc{Obs-Predictor}(\cdot)$ returns a sequence of stochastic reachable sets predicting the occupied regions by obstacles for $N_o$ steps ahead. The predictions can be obtained from a probabilistic or planning-based model \cite{rudenko2020human}. Given the predicted dynamic obstacle set $\hat{\mathcal{X}}^d_{o}(t)$ for each dynamic agent $d$, we derive the safe set $\hat{\mathcal{X}}^d_{s}(t)\subseteq\mathcal{X}\backslash \hat{\mathcal{X}}^d_{o}(t)$ with respect to the agent $d$ for $N_o$ steps ahead,
\begin{align}\label{safe-set-dynamic}
    \hat{\mathcal{X}}^d_{s}(t)= \{\mathbf{x}\!\in\! \mathcal{X}\;| h^d(\mathbf{x}(t))\geq 0 \},
\end{align}
where $t\in \{t_0+i T_s\;|\;i=0,1,\cdots,N_o\}$. The defined function $h^d(\mathbf{x})$ is then employed as a Control Barrier Function to obtain the control inputs that keep the robot state $\mathbf{x}$ in the safe region $\hat{\mathcal{X}}^d_s$ with respect to the dynamic agents $d=1,\cdots,D,$ for $N_o$ steps ahead. For a set of static obstacles $\mathcal{X}^c_o$, for each static obstacle $c=1,\cdots,C$, safety set $\mathcal{X}^c_{s}\subseteq\mathcal{X}\backslash \mathcal{X}^c_{o}$ can be defined as
\begin{align}\label{safe-set-static}
    \mathcal{X}^c_{s}= \{\mathbf{x}\in \mathcal{X}\;| h^c(\mathbf{x})\geq 0 \}.
\end{align}

Given the safety sets $\mathcal{X}^c_{s}$ and $\hat{\mathcal{X}}^d_{s}$, the algorithm expands a tree from the current position of robot $\mathbf{x}(t_0)$ and returns a partially grown tree $\mathcal{G}=(\mathcal{V},\mathcal{E})$ to choose the most promising path. To select such a path, algorithm first obtains the vertex with the minimum cost. The most promising path is the path from the root (the current position of robot) to the vertex with the minimum cost $z$ (defined in (\ref{cost2}) in Sec. \ref{SEC:safe-steering}). $\textsc{Extract-Control}(\cdot)$ query finds this path and extracts the first calculated control $\mathbf{u} (t_0)$ stored in the first edge segment of the selected partial path. $\textsc{Execute-Control}(\cdot)$ then sends $\mathbf{u} (t_0)$ to the robot for execution. In the following section, we explain how a safe control sequence is generated for each path segment. \

\begin{algorithm}[t]
\caption{$\textsc{CBF-TB-RRT}$}
\label{CBF-alg}
\begin{algorithmic}[1]
\renewcommand{\algorithmicrequire}{\textbf{Input:}}
\renewcommand{\algorithmicensure}{\textbf{Output:}}
\REQUIRE $\mathcal{X}_{g}$, $T_s$, $N_o$, $N_s$
\ENSURE  $\mathcal{G}=(\mathcal{V},\mathcal{E}),$ $\sigma$
\STATE $t_0\leftarrow \textsc{Clock}()$ 
\STATE $\mathbf{x}(t_0)\leftarrow \textsc{Robot-Pose}()$ 
\STATE $\sigma\leftarrow\mathbf{x}(t_0)$, $\mathcal{G}\leftarrow\emptyset$, $\mathcal{T}\leftarrow\emptyset$
\WHILE{$\mathbf{x}(t_0)\notin\mathcal{X}_{g}$}
\STATE $\mathcal{T}\leftarrow \textsc{Obs-Pose}()$
\STATE $\mathcal{X}^d_o(t)\leftarrow \textsc{Obs-Predictor}(\mathcal{T},N_o)$
\WHILE{$\textsc{clock}()<t_0+T_s$}
\STATE $\textsc{Grow}(\mathcal{G},\mathcal{X}^d_{o}(t),N_s)$
\ENDWHILE
\STATE $\mathbf{u} (t_0)\leftarrow \textsc{Extract-Control}(\mathcal{G})$
\STATE $\textsc{Execute-Control}(\mathbf{u} (t_0))$
\STATE $t_0\leftarrow \textsc{clock}()$
\STATE $\mathbf{x}(t_0)\leftarrow \textsc{Robot-Pose}()$ 
\STATE $\sigma\leftarrow\mathbf{x}(t_0)$
\ENDWHILE
\end{algorithmic}
\end{algorithm}
\begin{algorithm}[t]
\caption{$\textsc{Grow}$}
\label{grow-alg}
\begin{algorithmic}[1]
\renewcommand{\algorithmicrequire}{\textbf{Input:}}
\REQUIRE $\mathcal{G},\mathcal{X}_{o}(t),N_s$
\STATE $\mathcal{V}\leftarrow\emptyset$, $\mathcal{E}\leftarrow\emptyset$
\STATE $v_v\leftarrow \textsc{Vertex-Sample}(\mathcal{G},p_v)$
\STATE $\mathbf{x}_{rand}\leftarrow \textsc{State-Sample}(v_v,p_s)$
\STATE $\mathbf{u}_{ref}\leftarrow \textsc{Ref-Sample}(\mathbf{x}_{rand},p_u)$ 
\STATE $\mathbf{U}_{seg},\mathbf{X}_{seg},\mathbf{x}_{new},t_{new}\!\leftarrow\!\textsc{Steer}(\mathbf{x}_{rand},\mathbf{u}_{ref},N_s,v_v)$ 
\IF{$\mathbf{x}_{new}\neq \emptyset$}
\STATE $z_{new}\leftarrow\textsc{Cost-Calculator}(\mathbf{x}_{new})$
\STATE $\mathcal{V}\leftarrow(\mathbf{x}_{new},z_{new},t_{new}),$ $\mathcal{E}\leftarrow(\mathbf{X}_{seg},\mathbf{U}_{seg})$
\STATE $\mathcal{G}\leftarrow(\mathcal{V},\mathcal{E})$
\ENDIF
\end{algorithmic}
\end{algorithm}
%%%%%%%%%%%%%%%%%%%%%%%%%%%%%%%%%%%%%%%%%
\subsection{CBF-based Safe Steering Mechanism}\label{SEC:safe-steering}
To generate safe path segments at each tree expansion cycle $T_s$, we employed the safe steering method proposed in \cite{yang2019sampling}, and extended it for safe navigation in the presence of moving obstacles with unknown dynamics in a real-time application. Here, we present the safe steering method and then we discuss how we extend the technique. In CBF-based tree expansion, the state sampling mechanism is different than the classical RRT problem \cite{lavalle1998rapidly}. Depends on the task requirement or system dynamical constraints, some variables of the state vector $\mathbf{x}_v\in \mathcal{X}$ stored in a vertex $v_v\in\mathcal{V}$, may not play  an essential role in the given task. As an example, assume the robot is following the bicycle model with state vector $\mathbf{x}=[x ,y ,\theta ]^T$ that includes longitudinal and lateral positions, and heading angle, respectively. In such system, if the navigation task is to move from a position $[x(0),y(0)]^T$ at time $t=0s$ to the position $[x(2),y(2)]^T$ at time $t=2s$, the heading angle $\theta$ is not an important variable in this navigation task. We denote such variables as free state variables $x_{free}\in \mathbf{x}$. In state sampling, we perform the sampling over these free variables of the state $\mathbf{x}_v$. The collision avoidance is performed by steering the sampled state toward a safe region $\mathcal{X}_s$ with an optimization-based controller designed using the properties of CBF. To this end, the tree expansion is performed in the following steps,
\begin{enumerate}
    \item $\textsc{Vertex-Sample}(\cdot)$ samples $v_v$ over the existing vertices of $\mathcal{G}$ using a probability distribution $p_v$.
    \item $\textsc{State-Sample}(\cdot)$ extracts the state vector $\mathbf{x}_{v}\in\mathcal{X}$ stored in $v_v$ and updates the value of its free state variables $x_{free}$ using a probability distribution $p_s(x_{free})$. We denote the new randomly generated state by $\mathbf{x}_{rand}$. 
    \item $\textsc{Reference-Sample}(\cdot)$ samples over the action space $\mathcal{U}$ of the robot with a probability distribution $p_u$ if no free state variable exists in the sampled vertex $v_v$. The sampled input is denoted with $\mathbf{u}_{ref}$.
\end{enumerate}
Given the sampled vertex $v_v$, sampled state $\mathbf{x}_{rand}$, and the reference input $\mathbf{u}_{ref}$, $\textsc{Steer}(\cdot)$ solves a sequence of quadratic programmings (QP) to generate a sequence of control inputs $\mathbf{U}_{seg}$ subject to the CBF-based safety constraints derived from $\mathcal{X}_s(t)=\{\hat{\mathcal{X}}^d_s(t)\}\cap\{\mathcal{X}^c_s\}$ with respect to all dynamic obstacles $d=1,\cdots,D,$ and the static obstacles $c=1,\cdots,C$. The input sequence $\mathbf{U}_{seg}$ results a dynamically feasible and safe path segment $\mathbf{X}_{seg}$ starting from the vertex $v_v$ toward a safe state $\mathbf{x}_{new}\in\mathcal{X}_s(t)$. The quadratic programming (QP) based controller with CBF constraints guarantee the safety of generated control and path segment. The variable $N_s$ determines the length of each path segment (number of intermediate states in each segment), e.g., $N_s=3$ in Fig. \ref{fig:RRT}.\

Given the predicted safe set (\ref{safe-set-dynamic}) for the dynamic obstacles $d=1,\cdots,D$, and the safe set (\ref{safe-set-static}) for the static obstacles $c=1,\cdots,C$, we can use the properties of CBFs to generate a safe control input $\mathbf{u}$ by solving the following QP,
\begin{align} 
& \min_{\mathbf{u}\in \mathcal{U}} && J(\mathbf{u}) \label{cost1}\\ 
&\mbox{ s.t.} && L_f h^d(\mathbf{x}) +L_g h^d(\mathbf{x}) \mathbf{u}+\alpha(h^d(\mathbf{x}))\geq 0,\label{constraints-dyn}\\ 
&&& L_f h^c(\mathbf{x}) +L_g h^c(\mathbf{x}) \mathbf{u}+\alpha(h^c(\mathbf{x}))\geq 0,\label{constraints-stat}
% \\ 
% &&& \mathbf{u}\in \mathcal{U},
% \label{constraint-control}
\end{align}
The safety constraints (\ref{constraints-dyn}) and (\ref{constraints-stat}) are derived from  Ineq. (\ref{CBF}) for the dynamic obstacles $d=1,\cdots,D$, and the static obstacles $c=1,\cdots,C$, respectively. The objective function (\ref{cost1}) is defined as
\begin{align}\label{obj}
     J(\mathbf{u}) = (\mathbf{u}-\mathbf{u}_{ref})^T \mathbf{H} (\mathbf{u}-\mathbf{u}_{ref}),
\end{align}
where $\mathbf{H}$ is a diagonal matrix with non-negative elements. Note that in the above program, the objective function (\ref{obj}) and the constraints (\ref{constraints-dyn}) and (\ref{constraints-stat}) are quadratic and linear in the search parameter $\mathbf{u}$, respectively.\ 
Hence, given $\mathbf{u}_{ref}$, the defined QP problem can be solved starting from $\mathbf{x}_v$ for $N_s$ number of iterations to generate a safe control sequence $\mathbf{U}_{seg}$ and a path segment $\mathbf{X}_{seg}$ ending with $\mathbf{x}_{new}$. The cost of new vertex is then calculated in Alg. \ref{grow-alg} by $\textsc{Cost-Calculator}()$ query as follows
\begin{align}\label{cost2}
    z_{new}=\frac{a_1\text{dist}(\mathbf{x}_{new},\mathcal{X}_g)}{a_2 h(\mathbf{x}_{new})},
\end{align}
where $\text{dist}(\cdot)$ represents the distance between $\mathbf{x}_{new}$ and the goal set $\mathcal{X}_g$, and $a_1$ and $a_2$ are the weights of numerator and denominator terms, respectively. Here, $a_1$ and $a_2$ can introduce a trade-off between a highly conservative path and closeness to the goal set. \

In our proposed algorithm, both dynamic feasibility and safety of a generated path is guaranteed. Using real-time CBF-based navigation, we enable the robot to generate a set of safe path segments in presence of unknown dynamic obstacles and select the safest path while navigating toward the goal region. Moreover, with the combination of an optimization-based controller and a sampling-based motion planning technique, we employ the strength of sampling-based methods to explore the environment. We achieve this by exploring a feasible safe manoeuvre in the control and state spaces through sampling. In comparison with the proposed method in \cite{yang2019sampling}, we employed the CBF-based steering method in an online fashion combined with TB-RRT. Moreover, we predict the motion of dynamic obstacles through generating level-sets from a given prediction model without any information on the obstacle dynamics.\

\subsection{Implementation Details} \label{SEC:implement-details}
In our implementation, we consider the nonholonomic unicycle model for the robot dynamics as 
\begin{align}\label{rot-model}
    \dot{\mathbf{x}} (t)=\mathbf{g}(\mathbf{x}(t))\mathbf{u} (t)={\small\begin{bmatrix}
    \cos(\theta (t)) & 0\\
    \sin(\theta (t)) & 0\\
     0 & 1
    \end{bmatrix}}\mathbf{u} (t).
\end{align}
where states are $\mathbf{x}(t)\!=\![x (t),y (t),\theta (t)]^T\!\in\!\mathcal{X}\!\subseteq \mathbb{R}^2\!\times\![-\pi,\pi)$ and control input vector is $\mathbf{u} (t)=[v (t),\omega (t)]^T\in \mathcal{U}\subseteq \mathbb{R}^2$. The parameters $x (t)$, $y (t)$, $\theta(t)$ denote the longitudinal and lateral positions of the robot and heading angle, respectively. The controls $v (t)$ and $\omega(t)$ also represent the linear and angular velocities of the robot, respectively. The robot dynamics is integrated using the Euler scheme with the step size $T_s$. Moreover, the goal set $\mathcal{X}_{g}\subset\mathcal{X}$ of robot can describe a set of position states in $\mathbb{R}^2$ as follows
\begin{align}\label{G-set2}
    \mathcal{X}_g =: \big\{\mathbf{x} \in \mathcal{X}\;|\; \big\lVert [x,y]^T-\mathbf{x}_g\big\rVert^2_2 - r_g^2 \leq 0 \big\},   
\end{align}
where $\lVert\cdot\rVert_2$ denotes the Euclidean norm, $\mathbf{x}_g=[x_g,y_g]^T$ is the center, and $r_g$ is the radius of the goal set. \ 

In vertex sampling, we sample over the existing vertices of the generated tree $\mathcal{G}$ at each time cycle $T_s$ following a discrete uniform distribution $p_v=\frac{1}{|\mathcal{V}|}$. As the robot position at each vertex is fixed, the only remaining free variable is $\theta$. Given the heading toward goal $\theta_g\in[-\pi\times\pi)$ at each sample time, the following Gaussian distribution is used for sampling the free state $\theta$
\begin{align}\label{sigma}
    p_s(\theta)=\frac{1}{\sigma_{\theta}\sqrt{2\pi}}\text{exp}(-\frac{(\theta-\theta_g)^2}{2\sigma^2_{\theta}}),
\end{align}
where $\sigma_{\theta}$ is the variance that can determine the tree exploration range. Reference input $v_{ref}$ is selected randomly over the allowed range of speed with a uniform distribution, and the reference input $\omega_{ref}$ is selected as $\omega_{ref}=a_{\omega}(\theta-\theta_g)$, where $a_{\omega}$ is the weight of angular difference $(\theta-\theta_g)$.\

\subsection{Dynamic Obstacle Motion Prediction}
In this paper, we use the joint sampling Markov decision process (JS-MDP) goal-directed planning-based method proposed in \cite{rudenko2018joint} to predict human motions in real-time. Given a tracklet $\mathcal{T}$ that records the human trajectories, JS-MDP predicts the humans' trajectory for $N_o$ steps ahead by stochastic policy sampling over the discretized orientation-velocity action set $\mathcal{A}^d$ following a deterministic kinematic motion transition function. As a result, JS-MDP returns a set of probabilistic occupancy maps $\mathcal{L}=\{\mathcal{L}_d^t\}$ stores the predicted occupied regions by each human $d$ for $N_o$ steps ahead ($t=t_0,\cdots,t_0+N_o T_s$) for each goal. The probabilistic map $\mathcal{L}^t_d:\mathbb{R}^2\rightarrow [0,1]$ presents a mapping from a Cartesian coordinate $x$-$y$ to the probability of occupancy.  The overall complexity of JS-MDP is $\mathcal{O}(K^d(D|\mathcal{G}^d|+N_o(|\mathcal{A}^d|)))$ where $K^d$ and $\mathcal{G}^d$ are the number of sampled trajectories and the set of goals, respectively. Readers are referred to \cite{rudenko2018joint} for further details.\
 
\begin{figure}[t]
	\centering
	\includegraphics[scale=0.5]{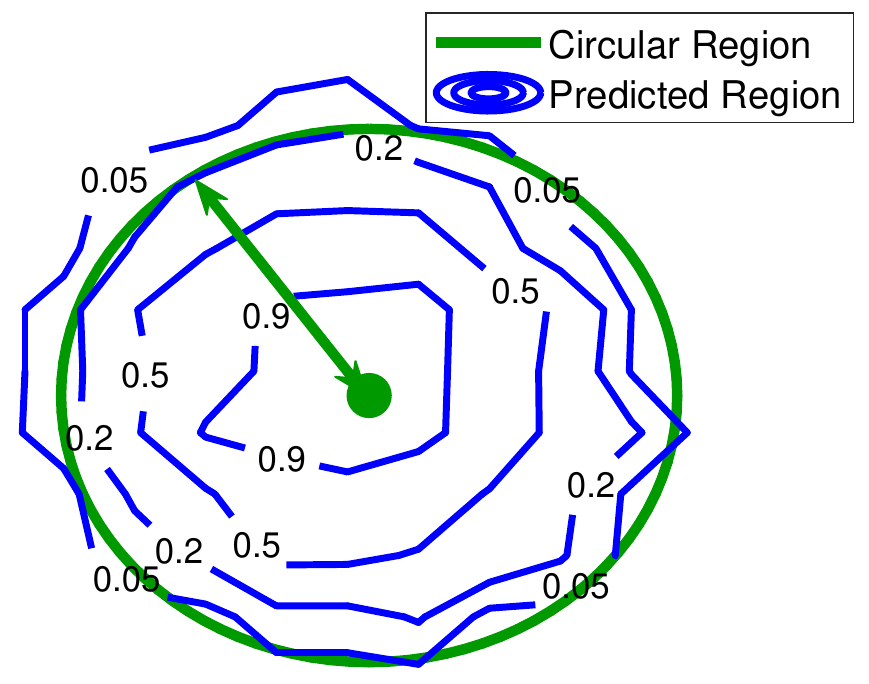}\vspace{-3pt}
	\caption{Level sets of the predicted occupied region by human with $p_o=0.2$.}
	\label{fig:circle_pack}
\vspace{-15pt}
\end{figure}

In our method, given the probabilistic occupancy map $\mathcal{L}_d^t$ for the agent $d$ at time $t$, we extract an ellipsoidal region from the occupied region in $x$-$y$ coordinates. The center of ellipsoid is the predicted occupied point in $\mathcal{L}_d^t$ with the maximum probability obtained as follows,
\begin{align}
     \hat{\mathbf{x}}_o^d(t)=\arg\max_{[x,y]^T}\mathcal{L}_d^t.
\end{align}
We also obtain the radius of ellipsoid to contain areas with probabilities greater than $p_o$ as
% \begin{align}
%      {\color{red} \hat{r}_o^d(t)=\max\big\lVert\mathcal{L}_d^t-\hat{\mathbf{x}}_o^d(t)\big\rVert_2\geq p_o.}
% \end{align}
\begin{align}\label{p_o}
      \hat{r}_o^d(t)=\max_{\{\mathbf{x}_o\;| \mathcal{L}_d^t(\mathbf{x}_o)\geq p_o\}}\big\lVert \mathbf{x}_o-\hat{\mathbf{x}}_o^d(t)\big\rVert_2.
\end{align}
Figure \ref{fig:circle_pack} shows an example of the level-sets. 

\begin{proposition}\label{prop1}
Given Assumption \ref{assum1}, the generated control by CBF-TB-RRT guarantees that the probability of collision at each time step is bounded by $p_o$.
\end{proposition}
\begin{remark}
The value $p_o$ determines how possible it is for the human to be in the extracted region and, therefore, enables us to capture the risk of collision with the human while generating path segments for the future during the tree expansion. JS-MDP is a goal-oriented method and does not assume the human's intent. As a result, the extracted predicted region does not necessarily increase in size with the length of horizon. Since the planning and control are performing in real-time for a finite horizon, this is an accurate assumption in practice. 
\end{remark}

 Given the predicted occupied region by the agent $d$, the continuously differentiable function $h^d(\mathbf{x})$ in (\ref{safe-set-dynamic}) is \begin{align}\label{safety-meas}
    h^d(\mathbf{x})=\big\lVert [x,y]^T-\hat{\mathbf{x}}_o^d(t)\big\rVert^2_2-\left(\hat{r}_o^d(t)+r_r\right)^2,
\end{align}
 where $r_r$ denotes the radius of robot. The safety measure $h^d(\mathbf{x})$ has a relative degree $1$ with respect to the input $v(t)$ and a relative degree $2$ with respect to the input $\omega(t)$. The consequence is that after deriving the safety constraint (\ref{constraints-dyn}), $\omega(t)$ does not appear in the constraint set since its corresponding multiplier in $L_g h^d(\mathbf{x})$ is zero. As in \cite{olfatinear,lindemann2020control}, to avoid this issue, we use a near-identity diffeomorphism to solve the problem for a closely related system. We use the following coordinate transformation 
\begin{equation}\label{coordinate transform}
    \Tilde{\mathbf{x}}(t)=\mathbf{x}(t)+\ell \mathbf{R}(\theta(t))\mathbf{e}_1,
\end{equation}
where $\ell>0$ is an arbitrary small constant, $~\mathbf{e}_1=[1~~0~~0]^T,~\mbox{and }\mathbf{R}(\theta(t))={\small\begin{bmatrix}
    \cos(\theta(t)) & -\sin(\theta(t)) & 0\\
    \sin(\theta(t)) & \cos(\theta(t)) & 0\\
    0 & 0 & 0
    \end{bmatrix}}.$
With this transformation, the model (\ref{rot-model}) is approximated to
\begin{align*}
    \dot{\Tilde{\mathbf{x}}}(t)=\Tilde{\mathbf{g}}(\Tilde{\mathbf{x}}(t))\mathbf{u}(t)={\small \begin{bmatrix}
    \cos(\theta(t)) & -\ell\sin(\theta(t))\\
    \sin(\theta(t)) & \ell\cos(\theta(t))\\
     0 & 1
    \end{bmatrix}}\begin{bmatrix}
    v(t)\\
    \omega(t)
    \end{bmatrix},
\end{align*}
where $\Tilde{\mathbf{x}}=[\Tilde{x},\Tilde{y},\Tilde{\theta}]^T$ is the approximated state vector and $\mathbf{u}$ is the original input. Since the maximum distance of $\mathbf{x}$ and $\Tilde{\mathbf{x}}$ is $\ell$, we modify the safety measure $(\ref{safety-meas})$ to account for this error as follows,
\begin{align}\label{safety-meas2}
    h^d(\Tilde{\mathbf{x}})=\big\lVert [\Tilde{x},\Tilde{y}]^T-\hat{\mathbf{x}}_o^d(t)\big\rVert^2_2-\left(\hat{r}_o^d(t)+\ell+r_r\right)^2.
\end{align}
Thus, the safety constraint (\ref{constraints-dyn}) is formulated as
\begin{align}\label{beta}
    L_f h^d(\Tilde{\mathbf{x}})+L_g h^d(\Tilde{\mathbf{x}}) \mathbf{u}+\beta h^d(\Tilde{\mathbf{x}})\geq 0
\end{align}
for each human agent $d$ where $\beta$ is a positive constant. We assumed the safety constraint ($\ref{constraints-stat}$) for each static obstacle $c=1,\cdots,C$, is also formulated in a similar format. \

\begin{figure*}[t!]
    \centering
    \begin{subfigure}[t]{0.33\textwidth}
        \centering
        \includegraphics[height=1.56in]{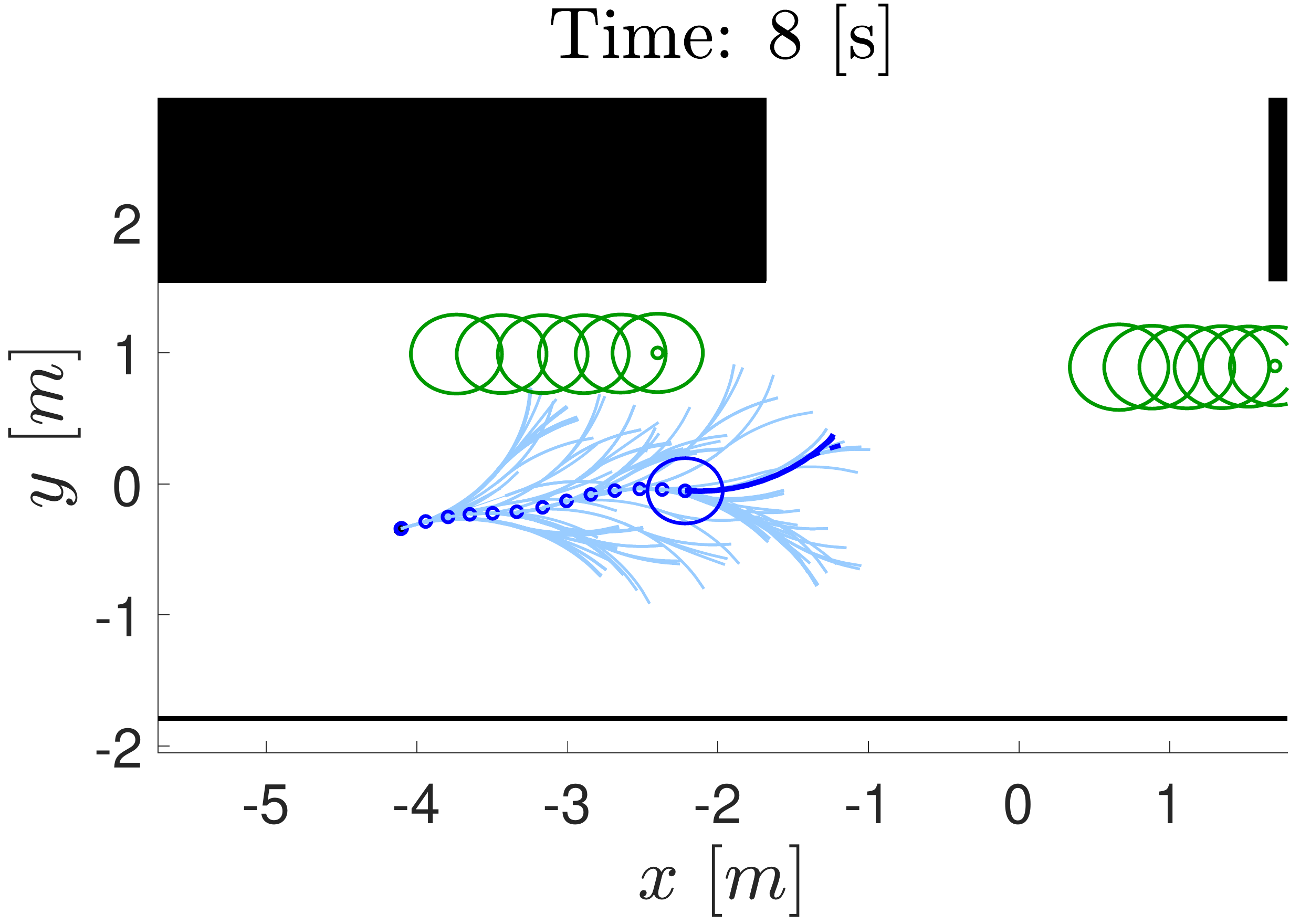}
        \caption{}
    \end{subfigure}%
    \begin{subfigure}[t]{0.33\textwidth}
        \centering
        \includegraphics[height=1.56in]{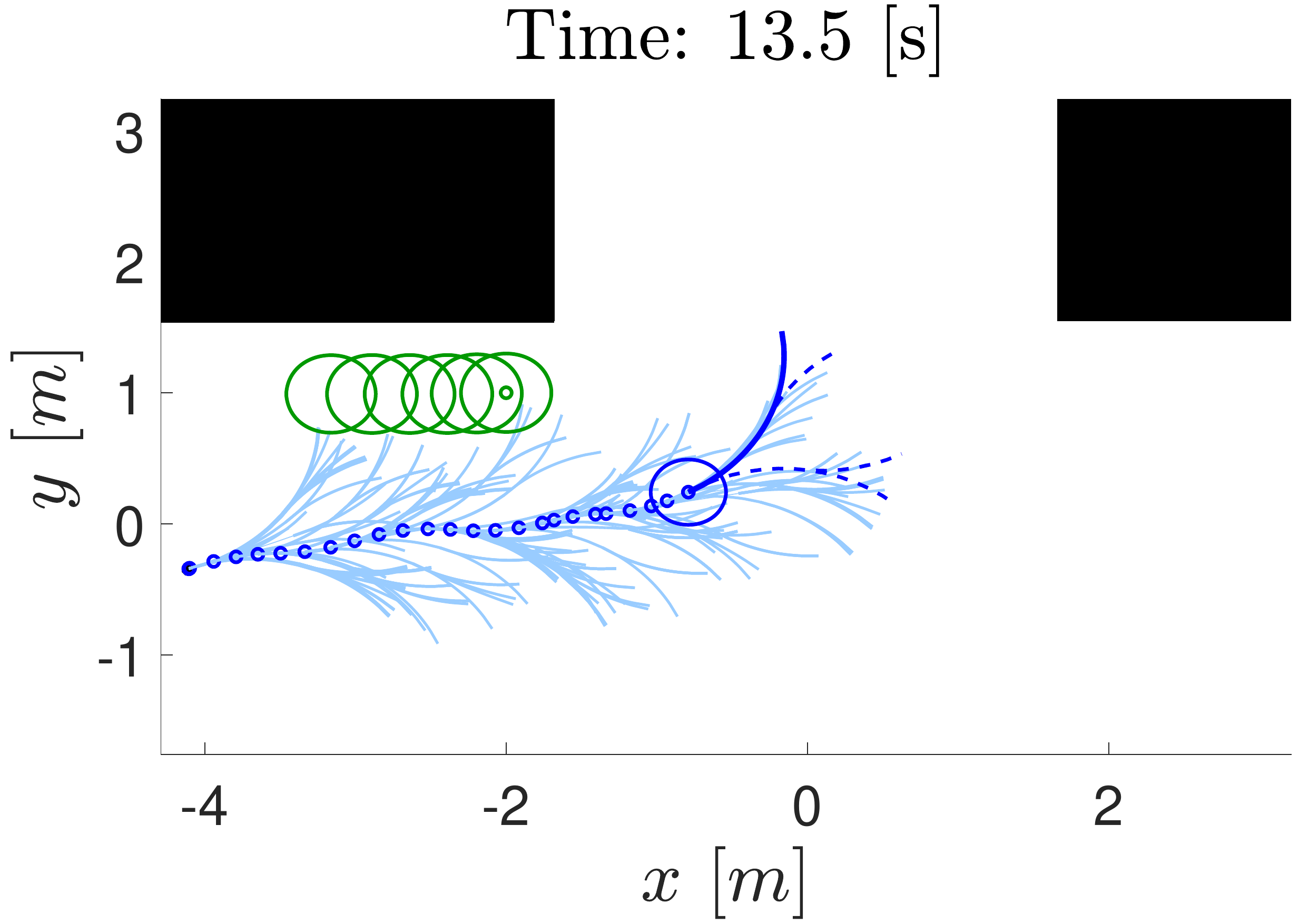}
        \caption{}
    \end{subfigure}
    \begin{subfigure}[t]{0.33\textwidth}
        \centering
        \includegraphics[height=1.5in]{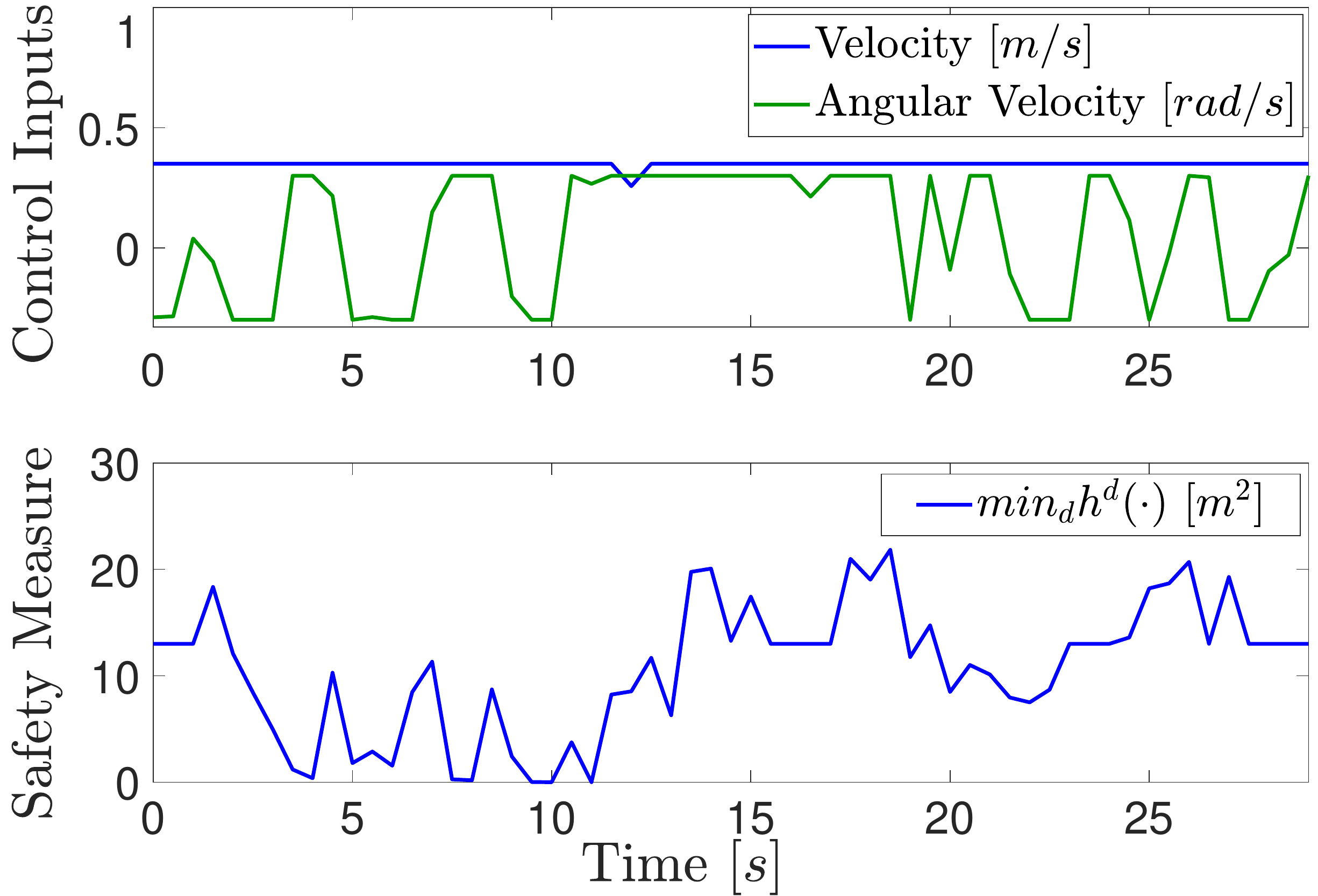}
        \caption{}
    \end{subfigure}
    \begin{subfigure}[t]{0.33\textwidth}
        \centering
        \includegraphics[height=1.56in]{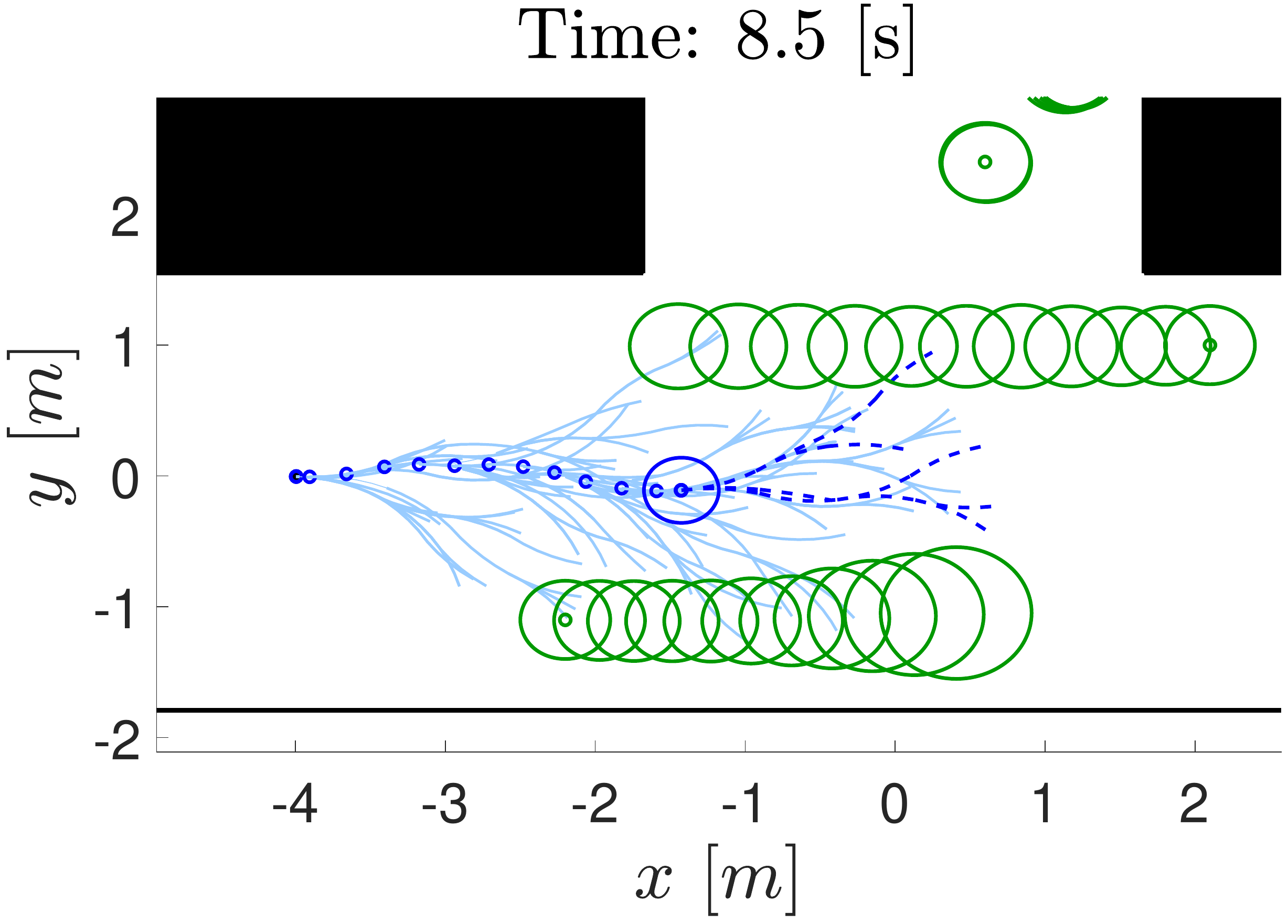}
        \caption{}
    \end{subfigure}%
    \begin{subfigure}[t]{0.33\textwidth}
        \centering
        \includegraphics[height=1.56in]{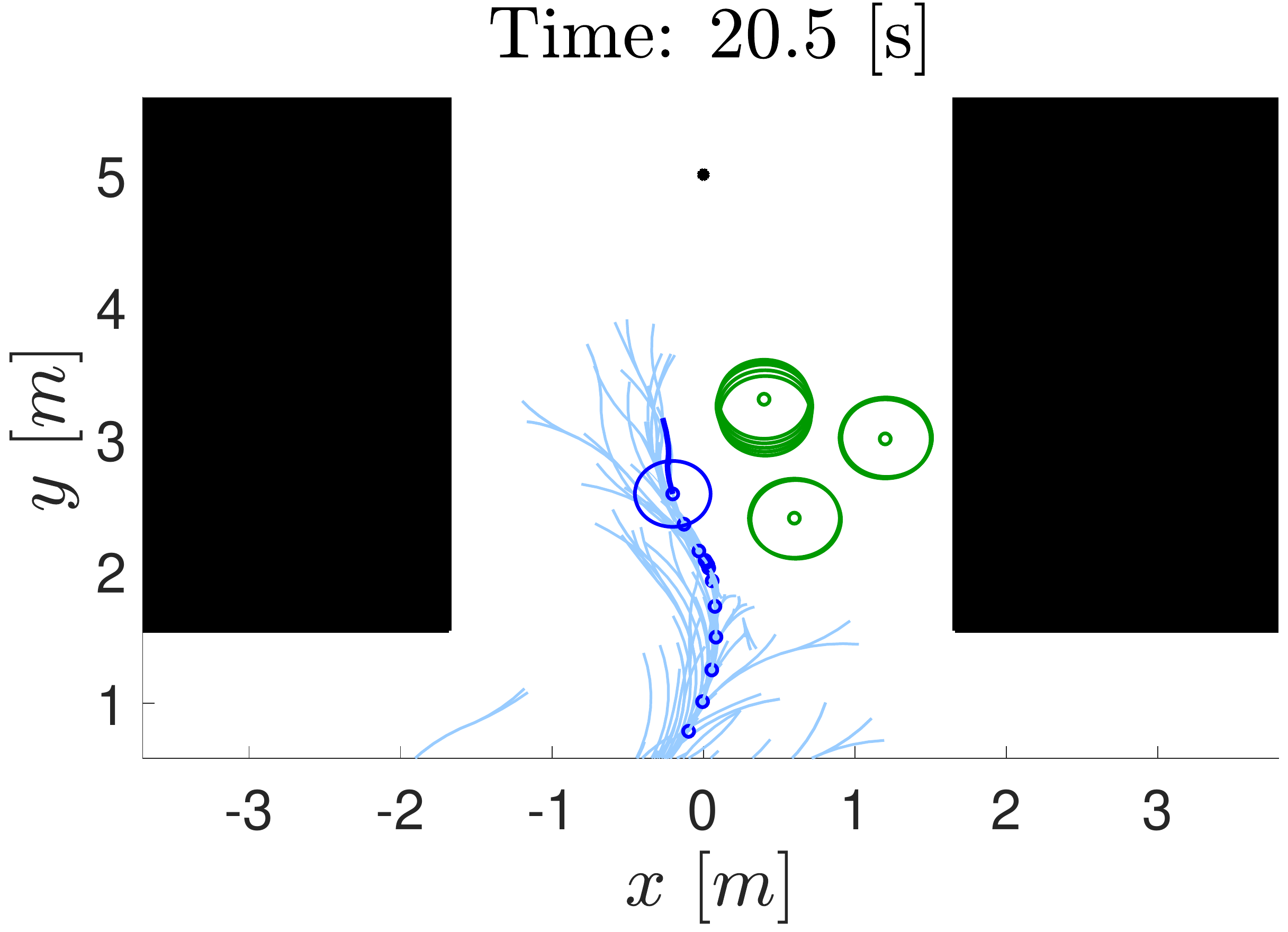}
        \caption{}
    \end{subfigure}
    \begin{subfigure}[t]{0.33\textwidth}
        \centering
        \includegraphics[height=1.5in]{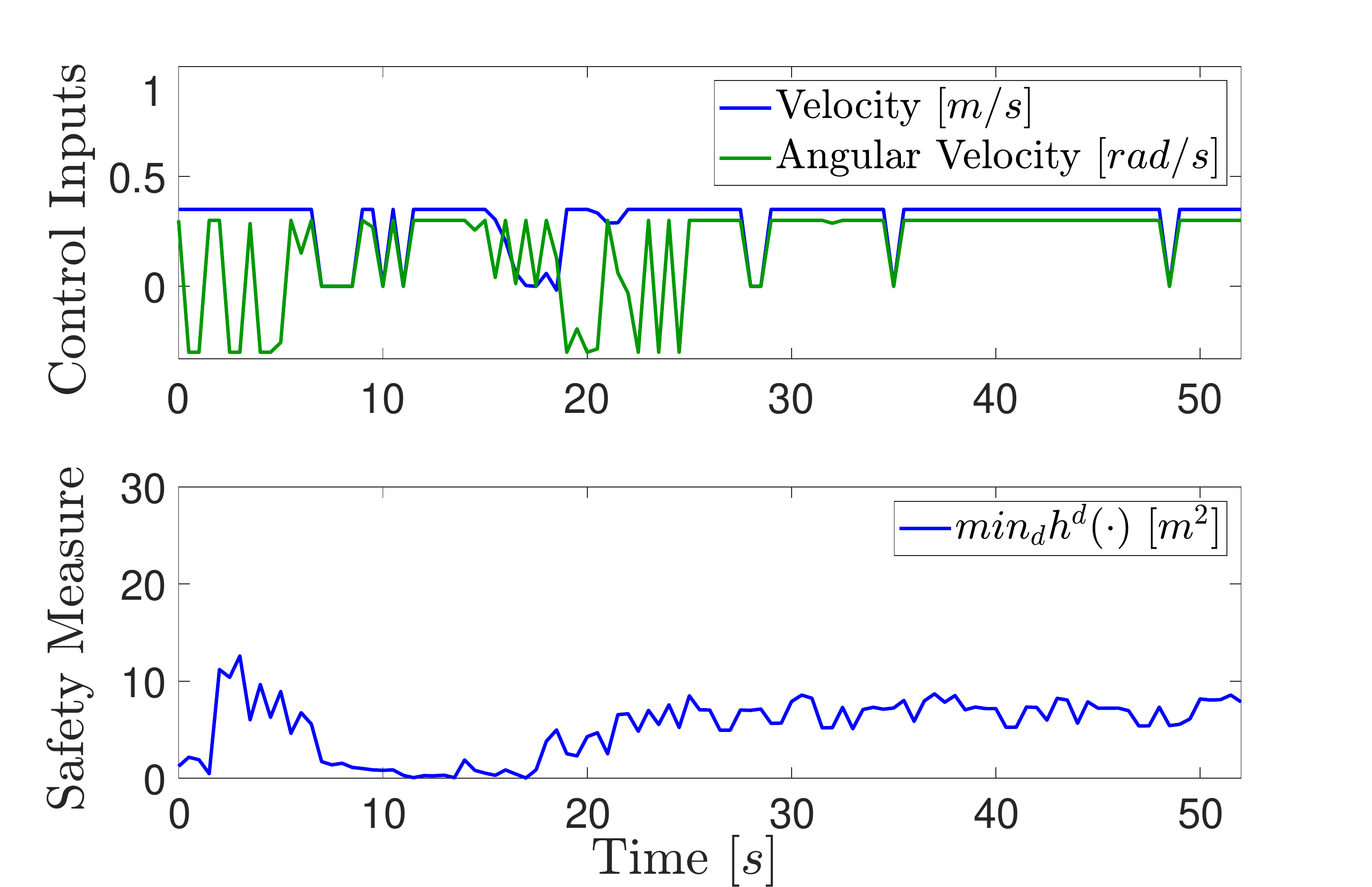}
        \caption{}
    \end{subfigure}
    \begin{subfigure}[t]{0.33\textwidth}
        \centering
        \includegraphics[height=1.56in]{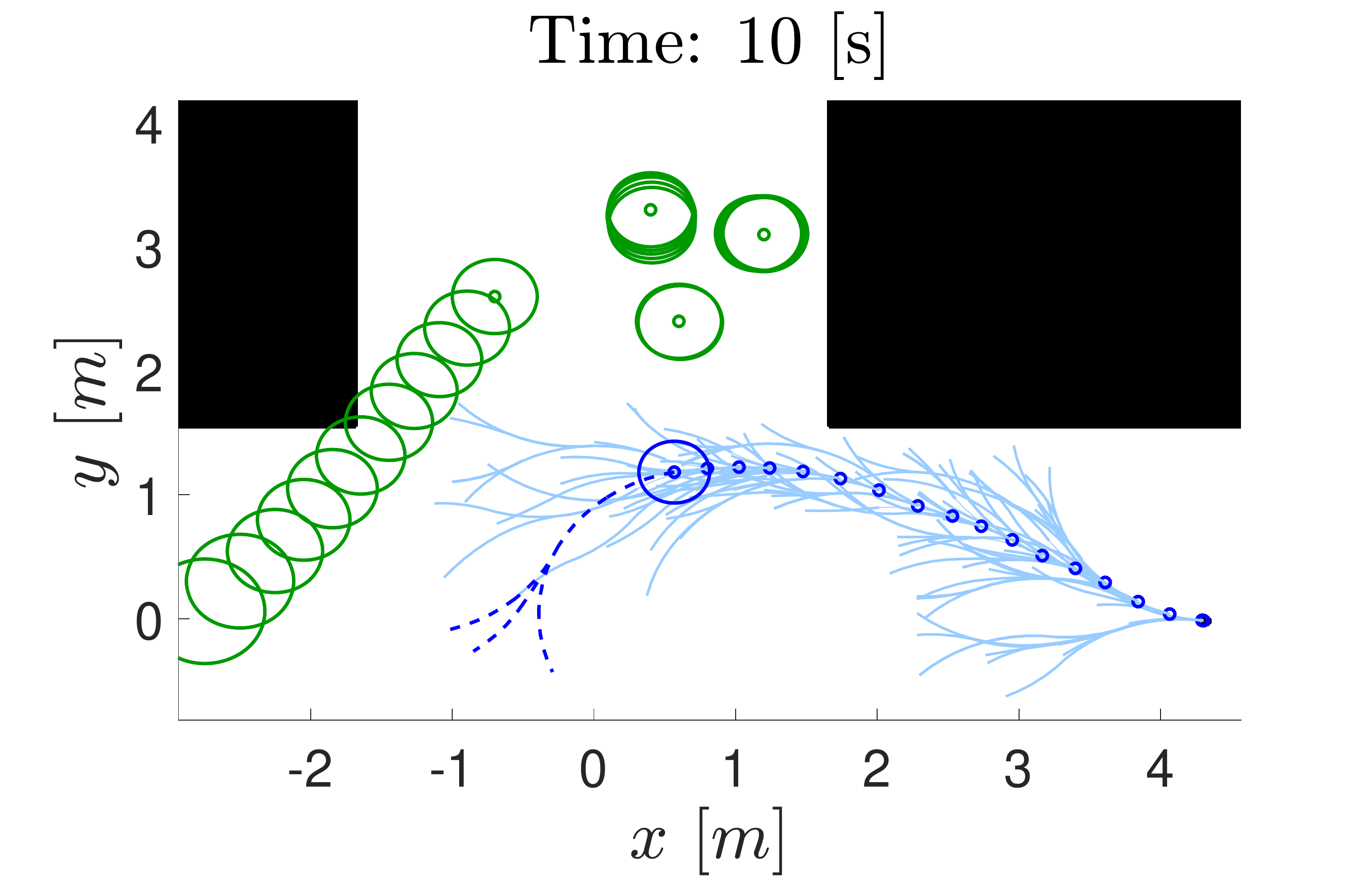}
        \caption{}
    \end{subfigure}%
    \begin{subfigure}[t]{0.33\textwidth}
        \centering
        \includegraphics[height=1.56in]{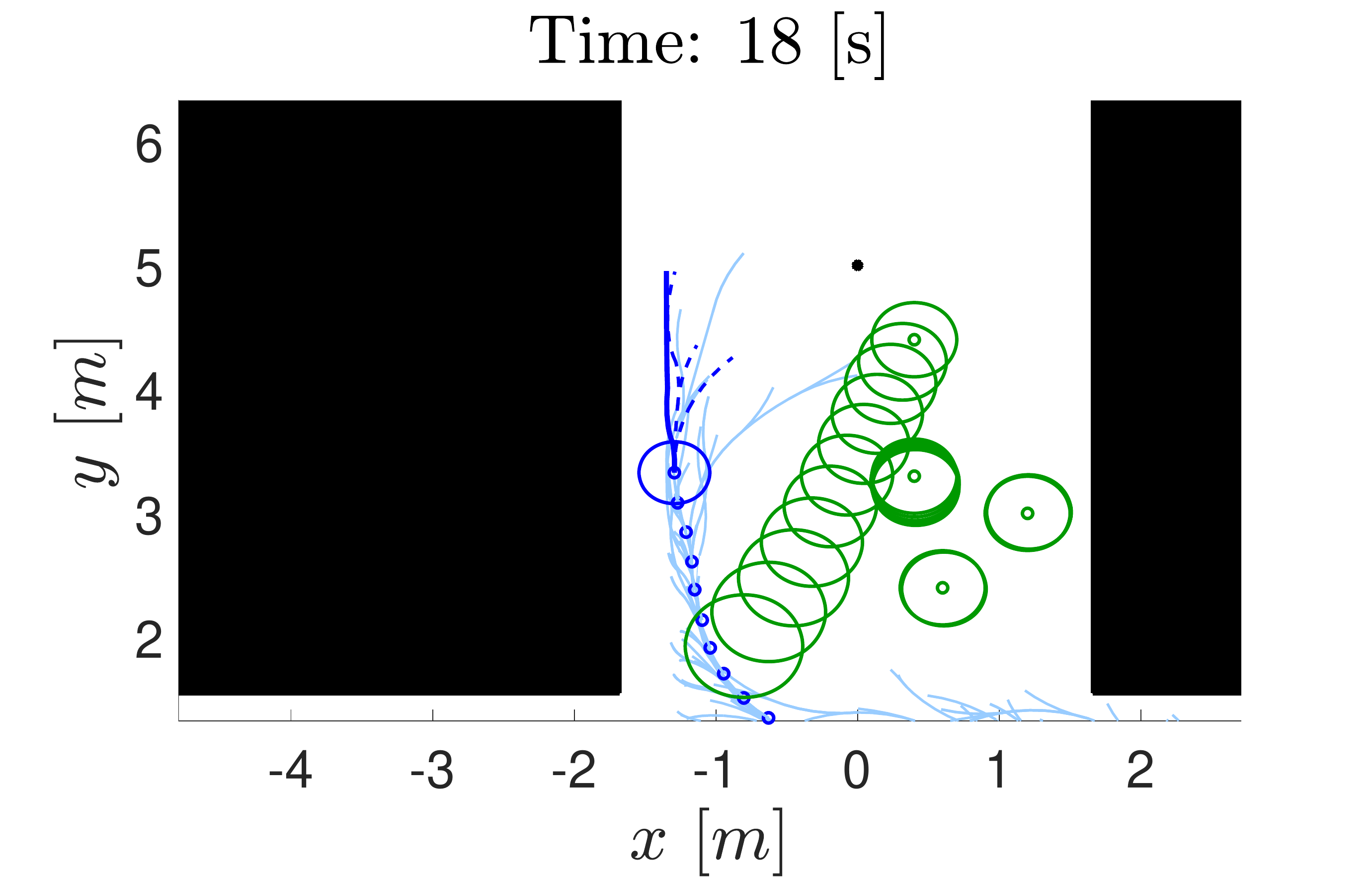}
        \caption{}
    \end{subfigure}
    \begin{subfigure}[t]{0.33\textwidth}
        \centering
        \includegraphics[height=1.5in]{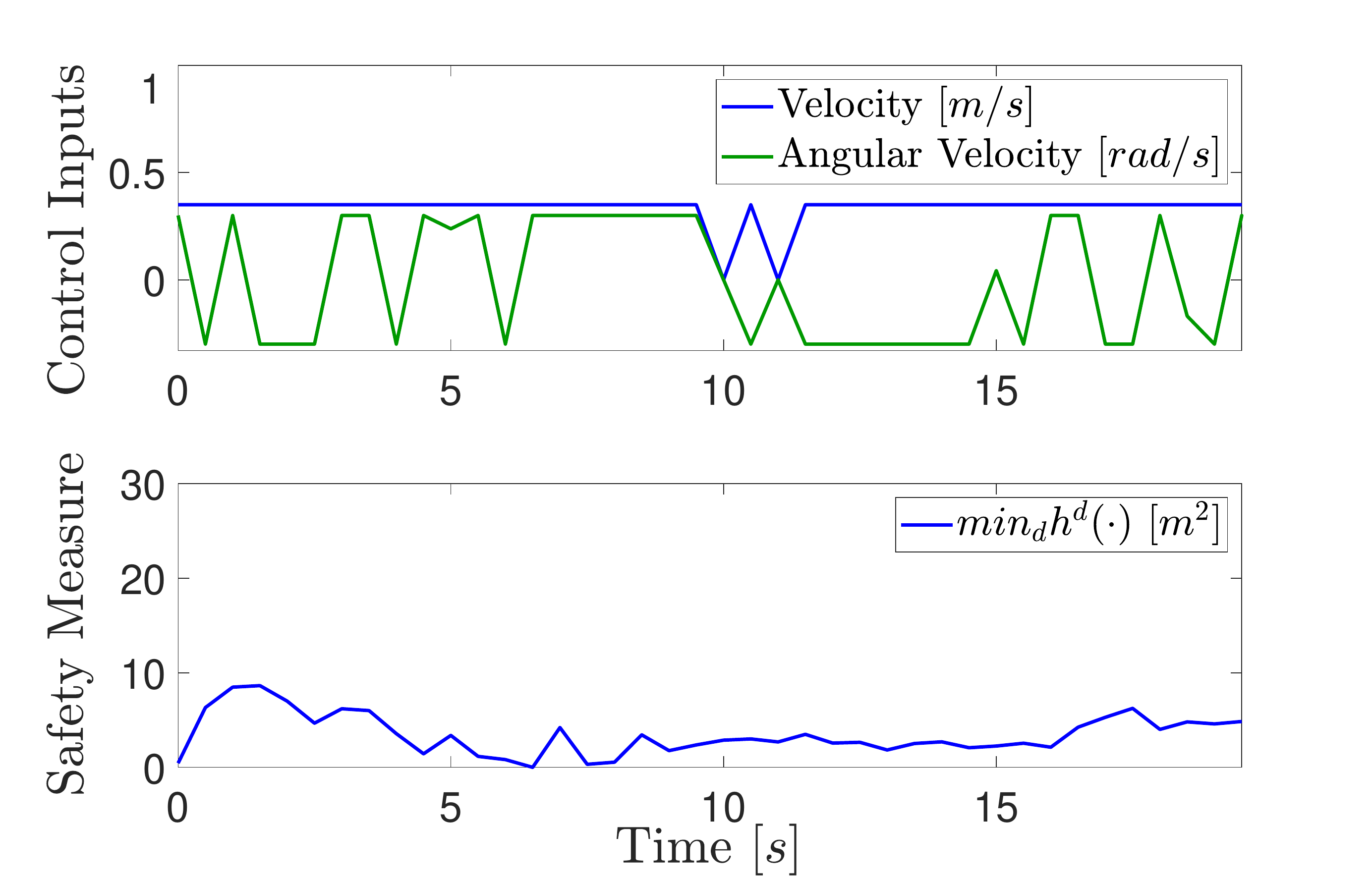}
        \caption{}
    \end{subfigure}
    \caption{The robot explores the environment to find a safe path toward goal using CBF-TB-RRT. At each frame, the expanded tree from $t=0$ to the current time is shown in light blue. Blue dots show the executed trajectory of the robot. Dark blue shows the expanded tree at the current time with the solid line representing the selected path. Green circles represent the future occupancy of dynamic obstacles with probability at least $20\%$. The generated controls and real-time safety measure $h(\Tilde{\mathbf{x}})$ demonstrated in the right-most figures (higher values are safer). Top row: a scenario with a group of humans moving behind each other. Middle and bottom rows: two complex scenarios with a group of static and dynamic agents in the scene.}
    \label{fig:RRT}
\end{figure*}

%%%%%%%%%%%%%%%%%%%%%%%%%%%%%%%%%%%%%%%%
%\subsection{Planning-based Approach for Human Motion Prediction}

%%%%%%%%%%%%%%%%%%%%%%%%%%%%%%%%%%%%%%%%
%%%%%%%%%%%%%%%%%%%%%%%%%%%%%%%%%%%%%%%%
%%%%%%%%%%%%%%%%%%%%%%%%%%%%%%%%%%%%%%%%
\section{SIMULATION RESULTS AND DISCUSSIONS}\label{sim-sec}

%%%%%%%%%%%%%%%%%%%%%%%%%%%%%%%%%%%%%%%%

%\subsection{Simulation Setup}
\begin{figure}[htbp]
\begin{minipage}{0.3\hsize}
  \includegraphics[width=\linewidth]{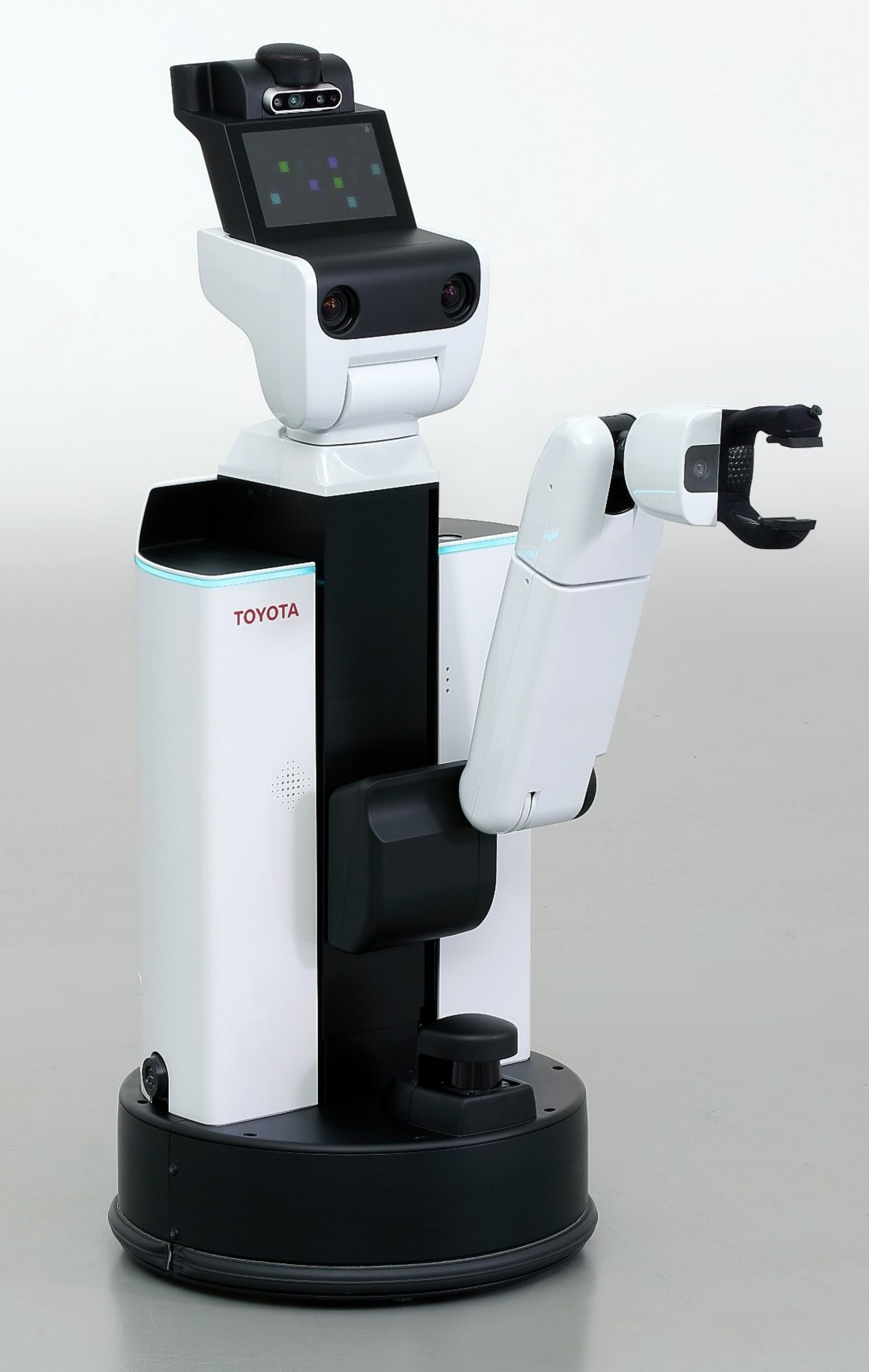}
  \caption{HSR}\vspace{-3pt}
  \label{fig:HSR}
\end{minipage}
\begin{minipage}{0.67\hsize}
  \makeatletter
  \def\@captype{table}
  \makeatother
  \caption{Parameters}
  \label{table:parameters}
  \small
  \centering
  \begin{tabular}{|l|l|}
    \hline
      $a_1$ (Eq. (\ref{cost2})) & $1$\\ \hline
      $a_2$ (Eq. (\ref{cost2})) & $1.5$ \\ \hline
      $\ell$ (Eq. (\ref{coordinate transform})) & $0.1$ \\ \hline
      $\beta $ (Eq. (\ref{beta})) & $100$\\ \hline
      $v_{max}$ (max velocity)& $0.33~[m/s]$ \\ \hline
      $\omega_{max}$ (max angular vel.)& $0.3~[rad/s]$\\ \hline
      $T_s$ (step size)& $0.1~[s]$ \\ \hline
      $p_o$ (Eq. (\ref{p_o}))& $0.2$\\ \hline
      $N_s$ (\# path segments)& $7$ \\ \hline
      $\sigma_{\theta} $ (Eq. (\ref{sigma}))& $1$ \\ \hline
      $r_r$ (Eq. (\ref{safety-meas2}))& $0.25~[m]$ \\ \hline
      $a_{\omega}$ (the weight of $\omega_{ref}$)& $0.2$ \\ \hline
  \end{tabular}
\end{minipage}
\vspace{-15pt}
\end{figure}

We evaluated the performance of CBF-TB-RRT in a high-fidelity HSR simulator (see Fig. \ref{fig:RRT}).
The HSR is a robot developed by Toyota Motor Corporation to operate in diverse environments while interacting with humans (Fig.~\ref{fig:HSR}).  It has a differential wheeled base and five degrees-of-freedom arm.
The HSR is 430~mm in diameter and 1,005~mm in height, the weight is 37~kg.
The simulator is based on ROS \cite{quigley2009ros} and Gazebo \cite{koenig2004design}. We used MATLAB\textsuperscript\textregistered~R2020a \cite{MATLAB:2020} to develop our algorithm, and ROS toolbox \cite{MatlabROS} for communication.\footnote{Code is available at \url{https://github.com/k1majd/CBF_TB_RRT.git}.}\

We designed three scenarios for the environment (in Fig. \ref{fig:gazebo}). In all scenarios, the robot should navigate through narrow corridors starting from the initial condition, $\mathbf{x}(0)=[-5,0,0]^T$ in the first two scenarios and $\mathbf{x}(0)=[5,0,\pi]^T$ in the third scenario, and move toward a goal region (\ref{G-set2}) with $\mathbf{x}_g=[0,5]^T$ and $r_g=0.3$. We name the corridor where the robot starts its movement as the \textit{start corridor} and the corridor where the goal is located as the \textit{goal corridor}. In our simulation setup, humans are assumed not to pay attention to their environment and they are following scripted paths with $1~[m/s]$ speed on average. In the first scenario, we considered two humans moving in the same direction behind each other to show how the robot generates a safe crossing trajectory when there is a short gap between the dynamic agents. In the second scenario, we considered a more complex case with a group of humans standing at the entrance of the goal corridor and two agents moving in opposite directions in the start corridor.  This final scenario is similar to the second scenario except a dynamic agent starts moving from the goal corridor, passes the group of standing agents, exits into the start corridor and turns right.
The weight matrix in (\ref{obj}) is $\mathbf{H}=Diag(10^5,10^5)$. The other parameters are shown in Table \ref{table:parameters}. During the tree expansion, at each state $\mathbf{x}$, we only considered the CBF constraints related to the agents that are in a distance $5$ or less from the robot to improve the efficiency. Moreover, the minimum value of sampling for linear velocity is assumed to be $0.2~[m/s]$. The reason is to prevent the robot from moving too slowly or stopping too often when there are no obstacles around. For demonstration purposes, the obstacle prediction horizons are chosen to be relatively small and relatively large values, $N_o = 5$ in the first scenario, $N_o=10$ in scenarios two and three, respectively. Choosing an optimal value for $N_o$ to balance the prediction horizon of the controller with the computational complexity will be studied in the future. 
\subsection{Results}
The simulation results are presented in Fig. \ref{fig:RRT}. The top, middle, and bottom figures correspond to the first, second, and third scenarios, respectively. For each scenario, the two left figures demonstrate two snapshots of the simulation alongside a figure on the right-hand side showing the robot's control inputs and the real-time minimum safety measure $h^{d}(\Tilde{\mathbf{x}})$ of the selected vertex over all moving agents. \

The first scenario (top row) evaluates the performance of our method with short-horizon obstacle predictions. As it can be seen in the first snapshot (Fig. \ref{fig:RRT} (a)), the robot does not find a safe gap to move toward the goal corridor. While the algorithm obtains a set of safe trajectories, robot selects the current position with zero velocity as the safest vertex considering the cost measure $z$. When the human clears the way, robot finds a sufficient gap to perform a safe motion toward goal (Fig. \ref{fig:RRT} (b)).\ 

The second scenario, evaluates the performance of our algorithm in a more complex case. In this scenario, the robot wants to proceed toward the goal region in the corridor on its left (goal corridor), but given the long-term prediction of human approaching, safety constraints guide the robot to move forward until it finds a better situation to turn and cross into the corridor (Fig. \ref{fig:RRT} (d)). In the second snapshot (Fig. \ref{fig:RRT} (e)), the robot finds the left-space of the goal corridor as the safest region to move and since the moving agent is still far, it turns and crosses safely. The expanded tree also leads the robot to move from the empty space of the goal corridor on the left to pass safely the group of humans standing on the right side of the corridor. Note that in this scenario, the standing humans are treated as dynamic obstacles but besides some small local movements, the predicted bounds show no motion.\

In the third scenario, after arriving to the goal corridor,  the group of standing humans and the moving agent leave no free space for the robot to proceed toward the goal region. Thus, the robot stops until it finds a free space to move (Fig. \ref{fig:RRT} (g)). In the second snapshot (Fig. \ref{fig:RRT} (h)), when the robot passes the group of humans, a dynamic agent enters the goal corridor. The safety constraints guide the robot to move closer to the wall and leave a free space for the dynamic agent to pass. The safety measure (Fig. \ref{fig:RRT} (i)) shows that while there is a narrow space for the robot to move, it finds a control that safely moves the robot toward the goal region while maintaining a safe distance with the moving and standing agents.  

\subsection{Discussion}

One advantage of integrating TB-RRT with CBF-based controller is to avoid the infeasibility of QP problem (\ref{cost1}). The sampling mechanism of RRT allows the robot to explore different directions for a safe feasible control. Besides the QP controller, the prediction horizon $N_o$, and the cost calculator weights $a_1$ and $a_2$, in (\ref{cost2}), affect the robot's motion behavior. Longer prediction horizon $N_o$ enables the robot to react more properly to the fast dynamic obstacles. In this case, robot leaves a free space for the human to pass as shown in the third scenario. However, long prediction horizon increases the computational time of JS-MDP. As mentioned above, choosing an optimal value for $N_o$ will be studied in our future work. While all the RRT expanded nodes at each iteration are safety guaranteed, the cost calculator weights $a_1$ and $a_2$ can determine how conservative the robot is. As shown in Fig. \ref{fig:RRT} (d), assigning larger values to $a_2$ rather than $a_1$ encourages the robot to select its current location that is a conservative strategy while there is a safe path segment closer to the goal. \

In general, the number of static and dynamic obstacles affect the computational cost of the QP controller and JD-MDP predictor. Indeed, the size of safety constraints (\ref{constraints-dyn}) and (\ref{constraints-stat}) increases with number of obstacles that affect the size of solution region (feasible region) in the QP problem. However, the computational time of the QP problem in our experiments is not affected by the number of obstacles. The JS-MDP computational cost grows with the number of dynamic obstacles, size of humans' goal set, and the number of samples. In our experiments, we improved the efficiency as follows. First, we only considered the CBF constraints related to the dynamic agents that are in a distance of $5~[m]$ or less of the robot. Second, we made a trade-off between the number of dynamic obstacles and the number of nodes in TB-RRT. In our prototype MATLAB implementation, the computation time of each prediction and control loop takes $0.6\pm0.1~[s]$ on average. We expect an implementation of the framework on C\textbackslash C++ will greatly improve the computational time that will be tested in our future work.
%In our experiments, the number of static and dynamic obstacles does not affect the computational complexity of the QP problem. Indeed, the size of safety constraints (\ref{constraints-dyn}) and (\ref{constraints-stat}) increase with number of obstacles that affects the size of solution region (feasible region) in the QP problem. However, the computational cost of the QP problem depends on the selected QP solver (we used QP solver of MATLAB) and, as mentioned, it is not affected by the number of obstacles in our experiments.

In this paper,  we demonstrate that our algorithms can safely handle scenarios where humans are inattentive and, hence, they do not react to the environment. Future work will address the problem where the humans react to the presence of robots. This adds a different level of complexity since an appropriate human movement model must be selected which takes into account the intention of the robot. That is, simple reactive human motion models would not be realistic enough unless they can capture the human-robot interaction. However, in our third scenario, we showed the robot can react fast and safely to the inattentive human who suddenly entered into the region where the robot is headed that is more challenging than a human who just reacts to the environment without any other cognitive process.\

Compared to the other variants of RRT, \cite{palmieri2016rrt, yang2019sampling, fulgenzi2010risk, wang2020eb, lavalle1998rapidly}, our method guarantees the safety of generated path segments without sampling control actions while considering the moving agents with unknown dynamics in real-time.

%%%%%%%%%%%%%%%%%%%%%%%%%%%%%%%%%%%%%%%%
%\subsection{Discussions}

%%%%%%%%%%%%%%%%%%%%%%%%%%%%%%%%%%%%%%%%
%%%%%%%%%%%%%%%%%%%%%%%%%%%%%%%%%%%%%%%%
\section{CONCLUSION AND FUTURE WORK}\vspace{-3pt}
In this paper, we provided a framework for safe motion planning in narrow corridors considering dynamic obstacles with unknown but predictable dynamics. We extended the TB-RRT algorithm by utilizing CBFs to ensure that all extensions of the tree satisfy safety requirements. To obtain the safe regions for a given prediction horizon, we used a  Joint Sampling MDP prediction model.
In our simulation results, we implemented our CBF-TB-RRT method on a Toyota HSR simulator. We showed three scenarios that the robot can safely navigate from an initial state toward a given goal through narrow corridors in presence of static and dynamic obstacles. 
In our future work, we aim to run our framework on a more realistic simulation setup with a reactive human motion model so as to test more challenging scenarios \cite{rudenko2020human}. We also plan to apply our method to a real HSR system.
%%%%%%%%%%%%%%%%%%%%%%%%%%%%%%%%%%%%%%%%
%%%%%%%%%%%%%%%%%%%%%%%%%%%%%%%%%%%%%%%%

\bibliographystyle{IEEEtran}
\bibliography{main.bib}

\end{document}